# Fault Prognosis of Turbofan Engines: Eventual Failure Prediction and Remaining Useful Life Estimation

Joseph Cohen[1], Xun Huan[2], and Jun Ni[2]

[1]*Michigan Institute for Data Science, University of Michigan, Ann Arbor, MI, 48109, United States of America*
*cohenyo@umich.edu*

[2]*Department of Mechanical Engineering, University of Michigan, Ann Arbor, MI, 48109, United States of America*
*xhuan@umich.edu*
*junni@umich.edu*

**ABSTRACT**

In the era of industrial big data, prognostics and health management is essential to improve the prediction of future failures to minimize inventory, maintenance, and human costs. Used for the 2021 PHM Data Challenge, the new Commercial Modular Aero-Propulsion System Simulation dataset from NASA is an open-source benchmark containing simulated turbofan engine units flown under realistic flight conditions. Deep learning approaches implemented previously for this application attempt to predict the remaining useful life of the engine units, but have not utilized labeled failure mode information, impeding practical usage and explainability. To address these limitations, a new prognostics approach is formulated with a customized loss function to simultaneously predict the current health state, the eventual failing component(s), and the remaining useful life. The proposed method incorporates principal component analysis to orthogonalize statistical time-domain features, which are inputs into supervised regressors such as random forests, extreme random forests, XGBoost, and artificial neural networks. The highest performing algorithm, ANN-Flux, achieves AUROC and AUPR scores exceeding 0.95 for each classification. In addition, ANN-Flux reduces the remaining useful life RMSE by 38% for the same test split of the dataset compared to past work, with significantly less computational cost.

## 1. INTRODUCTION

The field of prognostics and health management (PHM) has attracted recent research attention for large-scale, high-dimensional, and dynamic engineering systems. Typically performed on the component level, the goal of intelligent prognostic approaches is to predict the progression of degradation in advance to facilitate swift and responsible decision-making before catastrophic failure (Lee et al., 2014; Tsui et al., 2015). Typical PHM applications include data-driven fault diagnosis and prognosis of bearing failures (Shao et al., 2018) and gearbox failures (Li et al., 2016) utilizing vibration, current, and/or acoustic emission signals. As described by Liao and Köttig in (Liao & Köttig, 2014), PHM approaches can be separated into a few categories: physics-based, expert knowledge-based, or data-driven, with significant potential for hybridization. PHM is essential for reliable operation of safety-critical systems such as nuclear power plants, which have devastating consequences should catastrophic failures occur and are often difficult to predict due to the lack of historical, labeled failure data (Coble et al., 2015).

In light of the need for active PHM research, the PHM Society hosts annual data challenges open to the public to enable benchmarking for relevant industrial applications. This paper will focus on the 2021 PHM Data Challenge, which centered on accurately estimating the remaining useful life (RUL) for a small fleet of turbofan engines (M. A. Chao et al., 2021). The new Commercial Modular Aero-Propulsion System Simulation (N-CMAPSS) dataset, openly available in the NASA Prognostics Center of Excellence (PCoE) Data Set Repository (M. A. Chao et al., 2021), consists of synthetic run-to-failure trajectories operating under realistic flight conditions. The flight trajectories are generated using the C-MAPSS dynamical model from NASA Ames Research Center in collaboration with ETH Zurich and PARC (M. Chao et al., 2021). The turbofan engines experience 7 possible failure modes that involve efficiency and/or flow failures of 5 rotating subcomponents: fan, low-pressure compressor (LPC), high-pressure compressor (HPC), low-pressure turbine (LPT), and high-pressure turbine (HPT). A schematic representation of a turbofan engine unit is shown in Figure 1.

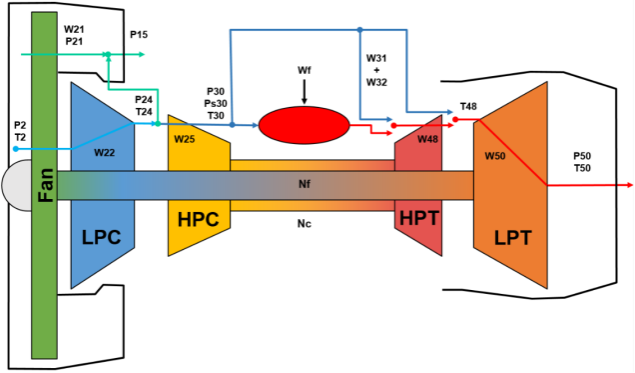

Figure 1. Turbofan engine schematic, courtesy of NASA Prognostics Center of Excellence (M. A. Chao et al., 2021)

The winning approaches of the 2021 PHM Data Challenge utilized deep learning techniques for RUL estimates. In (Lövberg, 2021), Lövberg proposed a neural network-based normalization procedure to effectively denoise the sensor measurements with respect to the dynamic flight conditions. After normalization, the input trajectories are passed into a deep convolutional neural network (CNN) with dilated convolutions in an approach that allows for variable input sequence lengths. Lövberg relied on the provided health state label to sample degraded sequences in their RUL prediction model. DeVol *et al.* proposed in (DeVol et al., 2021) the integration of inception modules within a CNN architecture to handle the variable trajectory lengths. DeVol *et al.* reported RUL prediction results using NASA's training-testing split in the N-CMAPSS dataset, streamlining reproducibility and benchmarking for this challenge problem. Finally, Solís-Martín *et al.* approached this problem by stacking two CNNs in sequence: an encoder model first used for dimensionality reduction and feature extraction, and a secondary model used for RUL prediction (Solís-Martín et al., 2021). Solís-Martín *et al.* used Bayesian hyperparameter optimization to tune their models and noted that their prediction results could be improved by reducing overfitting.

These approaches benefit from the strengths of deep learning; namely, they allow for feature representations to be learned via CNN rather than manually defined. Clever variations of CNNs, such as Lövberg's approach implementing dilated convolutions (Lövberg, 2021) and DeVol *et al.*'s usage of inception modules (DeVol et al., 2021), have allowed for accurate RUL estimation given varying flight trajectories and input lengths. However, there are key limitations to these approaches that inhibit their potential for practical use outside of the 2021 PHM Data Challenge. By focusing solely on RUL prediction, prior methods do not provide a holistic prognosis that predicts the eventual failing component(s). DeVol *et al.* mentioned that the resulting RUL predictions lack explainability, and that future work should utilize the labeled failure modes and components provided in the N-CMAPSS dataset to provide a more complete prognosis for turbofan engines (DeVol et al., 2021).

Our work significantly expands upon past efforts by broadening the research scope to encompass eventual failure prediction. Being able to accurately predict and isolate the reason for failure has important implications on maintenance decision-making, equipping operators with the capability to dispatch the appropriate experts and resources in a timely manner. Such predictive maintenance strategies can enable intelligent inventory optimization (Bousdekis et al., 2017) and reduce reactive maintenance costs, which may account for up to 40% of the overall budget in large industries (Bagavathiappan et al., 2013). To maximize applicability for a real-world scenario, we aim to simultaneously learn to predict three meaningful indicators: 1) the current health state; 2) the eventual failing component(s); and 3) the RUL until catastrophic failure. To the authors' knowledge, this is the first attempt at a unified model to effectively accomplish fault detection, isolation, and RUL estimation for the N-CMAPSS benchmark dataset. We accomplish these goals by first simplifying the feature extraction process to enable comparisons amongst state-of-the-art machine learning regressors. Then, we derive and optimize a specialized loss function that balances classification and regression objectives. We also compare the performance of state-of-the-art machine learning regressors and important pre-processing steps such as orthogonalization via principal components analysis (PCA). Our main contributions for this research effort are summarized as follows:

- Reformulating and expanding upon the 2021 PHM Data Challenge to include health state detection and forecasting eventual failures;
- Deriving a customized loss function to simultaneously optimize classification and regression PHM objectives;
- Accurately predicting health state, eventual failures, and RUL, with state-of-the-art regression approaches benchmarked with prior work.

In the following sections of the paper, we will detail our proposed methodology, our reproducible results, and provide comparisons to previous work.

## 2. METHODS

First, we will describe the N-CMAPSS dataset in detail, introducing the input variables and dataset composition in detail. With our expanded goal to predict the current health state and eventual failing components in addition to RUL, our proposed methodology encompasses both classification and regression objectives. Our method is summarized in three steps: 1) feature extraction; 2) feature standardization and orthogonalization via PCA; and 3) training a supervised machine learning model to obtain the final predictions. Figure 2 illustrates the proposed methodology.

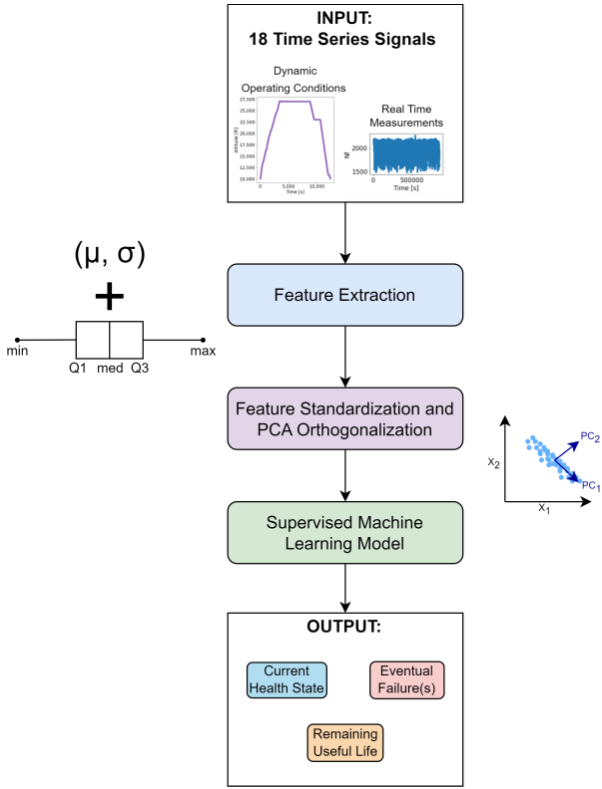

Figure 2. Proposed methodology flow for failure prediction

**2.1. Dataset Description**

The N-CMAPSS dataset consists of 8 provided subsets and contains 90 engine units in total. In our research, we combine flow and efficiency failures into one general failure category for each mechanical component. Table 1 provides a summary of the failure modes present in each subset. In the dataset, engine units have a lifetime rated typically between 60 and 100 cycles, with the overall objective being to estimate the RUL until catastrophic failure. Each flight cycle is of variable length and is characterized by 18 time series signals: 4 flight data descriptors $W = \{W_1, W_2, W_3, W_4\}$ summarizing the dynamic operating conditions, and 14 real-time sensor measurements $X_s = \{X_{s_1}, X_{s_2}, \ldots, X_{s_{14}}\}$. In addition to the time series signals, each cycle also includes auxiliary variables $A = \{A_1, A_2, A_3, A_4\}$ useful for understanding the context of a flight cycle: the unit number, cycle number, a categorical flight class variable $F_c$ representing the length of the flight (set to 1 for short flights, 2 for medium flights, and 3 for long flights), as well as a binary health state variable $h_s$ (set to 1 for healthy status and 0 for unhealthy status). We note that the simulated engines are flown past unhealthy operation until end of life (i.e., catastrophic failure). Table 2 provides a summary of the variables provided in the dataset.

In all, there are a total of 6825 flight cycles in the dataset, with engines averaging approximately 75 cycles per unit.

| Subset Name | Units | Failure Mode | Fan Fail | LPC Fail | HPC Fail | HPT Fail | LPT Fail |
|---|---|---|---|---|---|---|---|
| DS01 | 10 | 1 | No | No | No | Yes | No |
| DS03 | 15 | 2 | No | No | No | Yes | Yes |
| DS04 | 10 | 3 | Yes | No | No | No | No |
| DS05 | 10 | 4 | No | No | Yes | No | No |
| DS06 | 10 | 5 | No | Yes | Yes | No | No |
| DS07 | 10 | 6 | No | No | No | No | Yes |
| DS08a | 15 | 7 | Yes | Yes | Yes | Yes | Yes |
| DS08c | 15 | 7 | Yes | Yes | Yes | Yes | Yes |

Table 1. Failure mode description from the N-CMAPSS dataset (M. A. Chao et al., 2021)

**2.2. Feature Extraction**

Feature selection and extraction are necessary to reduce the input dimensionality of the dataset. Although there are only 90 turbofan engine units in the N-CMAPSS dataset as per Table 1, the dataset contains over 63 million timestamps and requires reduction for subsequent data processing. As in previous work, we aim to make predictions on a per-cycle basis (DeVol et al., 2021).

In this study, we extract cycle-wide statistical time domain features to summarize the distribution for each time series. These features include mean, standard deviation, and the five-number summary (minimum, first quartile, median, third quartile, and maximum) for all signals. We also extract features that are held constant per cycle such as the time duration of the cycle, the current cycle number, and the flight class $F_c$. In general terms, this feature selection and extraction method is applied for $n$ training cycles, with $x_j \in \mathbb{R}^n$ representing the vector of samples for the $j^{\text{th}}$ feature. Finally, the feature vectors are concatenated into a single data matrix containing all $p$ features, $X \in \mathbb{R}^{n \times p}$.

**2.3. Feature Standardization and PCA Orthogonalization**

Features extracted from the time series signals may be of different scales and units. As a result, standardization helps ensure that predictions are not influenced by these differences. First, we apply a min-max normalization scheme across all features to map all features in the bounded range [0, 1] as shown in Eq. (1):

$$\bar{x}_j = \frac{x_j - \min(x_j)}{\max(x_j) - \min(x_j)} \quad (1)$$

| Variable | Symbol | Description | Units |
|---|---|---|---|
| $A_1$ | unit | Unit number | - |
| $A_2$ | cycle | Flight cycle number | - |
| $A_3$ | $F_c$ | Flight class | - |
| $A_4$ | $h_s$ | Health state | - |
| $W_1$ | alt | Altitude | ft |
| $W_2$ | Mach | Flight Mach number | - |
| $W_3$ | TRA | Throttle-resolver angle | % |
| $W_4$ | T2 | Total temp. at fan inlet | °R |
| $Xs_1$ | Wf | Fuel flow | pps |
| $Xs_2$ | Nf | Physical fan speed | rpm |
| $Xs_3$ | Nc | Physical core speed | rpm |
| $Xs_4$ | T24 | Total temp. at LPC outlet | °R |
| $Xs_5$ | T30 | Total temp. at HPC outlet | °R |
| $Xs_6$ | T48 | Total temp. at HPT outlet | °R |
| $Xs_7$ | T50 | Total temp. at LPT outlet | °R |
| $Xs_8$ | P15 | Total pressure in bypass-duct | psia |
| $Xs_9$ | P2 | Total pressure at fan inlet | psia |
| $Xs_{10}$ | P21 | Total pressure at fan outlet | psia |
| $Xs_{11}$ | P24 | Total pressure at LPC outlet | psia |
| $Xs_{12}$ | Ps30 | Static pressure at HPC outlet | psia |
| $Xs_{13}$ | P40 | Total pressure at burner outlet | psia |
| $Xs_{14}$ | P50 | Total pressure at LPT outlet | psia |

Table 2. Auxiliary, flight descriptors, and sensor measurement variables used in N-CMAPSS dataset (M. A. Chao et al., 2021)

Once again, we concatenate the feature vectors into a normalized data matrix, $\overline{X} \in \mathbb{R}^{n \times p}$. After obtaining the normalized data matrix, PCA orthogonalization is recommended as a multivariate preprocessing step to obtain a set of uncorrelated variables. PCA is typically used to achieve dimensionality reduction by retaining the most important principal components (PCs) such that the explained variance is maximized (Jollife & Cadima, 2016). However, we have found that in practice, there is utility to keeping all PCs to improve training results. This is potentially because the features extracted are significantly correlated, and therefore simply using PCA for its orthogonalization benefits may improve the performance of gradient descent-based optimization methods employed in training. In Section 3, we will compare results with and without PCA orthogonalization for all models. PCA can be formulated as a linear transformation using the eigendecomposition of the sample correlation matrix $Q \in \mathbb{R}^{p \times p}$ of the features from $\overline{X}$, as shown in Eqs. (2)-(3):

$$Q = V\Lambda V^{\mathrm{T}} = [v_1, v_2, \ldots, v_p]\begin{bmatrix} \lambda_1 & \cdots & 0 \\ \vdots & \ddots & \vdots \\ 0 & \cdots & \lambda_p \end{bmatrix}[v_1, v_2, \ldots, v_p]^{\mathrm{T}} \quad (2)$$

$$\widetilde{X} = \overline{X}V = [\overline{x}_1, \overline{x}_2, \ldots, \overline{x}_p][v_1, v_2, \ldots, v_p] \quad (3)$$

where $v_1, v_2, \ldots, v_p$ are the principal components with corresponding eigenvalues $\lambda_1 \geq \lambda_2 \geq \cdots \geq \lambda_p \geq 0$. The resulting matrix $\widetilde{X} \in \mathbb{R}^{n \times p}$ is the newly orthogonalized training dataset scored along the PC axes.

### 2.4. Output Labeling Scheme

While the 2021 PHM Data Challenge formulation provides $h_s$ and failure mode information as possible inputs for a RUL prediction model, we aim to instead predict them as outputs encoded as binary variables. As mentioned previously, these additional outputs will provide a more comprehensive prognosis of the degraded turbofan engine unit. With these new outputs, we require a labeling scheme for training a model. For learning the current cycle health state $h_s$, we borrow the labels provided in the N-CMAPSS dataset, i.e., a label of "1" for healthy operation and "0" for unhealthy operation. We introduce a vector of possible eventual failures $y_{EF} = [y_{Fan}, y_{LPC}, y_{HPC}, y_{HPT}, y_{LPT}]^{\mathrm{T}}$, in which each variable $y_{comp} \in y_{EF}$ is binary, with the positive label indicating eventual failure as specified in Table 1. For example, for the DS06 subset in which the LPC and HPC components eventually fail, $y_{EF} = [0, 1, 1, 0, 0]^{\mathrm{T}}$. Lastly, for the RUL training label, we follow the N-CMAPSS convention, which provides $RUL \in \mathbb{Z}^*$ as calculated by subtracting the current cycle number from the total lifetime of the engine unit, i.e., $RUL = t_{EOL} - A_2$. With these definitions, we can prepare the ground truth vector of labels $y = [h_s, y_{EF}^{\mathrm{T}}, RUL]^{\mathrm{T}}$ paired with the features of each cycle.

### 2.5. Training Loss Function and Model Evaluation

Handling classification and regression objectives simultaneously provides additional complexity for training a predictive machine learning model. We propose optimizing a customized loss function that explicitly weighs both objectives. First, we base the RUL loss contribution from NASA's scoring criteria (M. A. Chao et al., 2021), which penalizes overestimation of RUL to favor conservative predictions and is defined in Eqs. (4)-(6):

$$s_c(RUL, \widehat{RUL}) = \frac{1}{n}\sum_{i=1}^{n} \exp(\alpha|RUL_i - \widehat{RUL}_i|) - 1 \quad (4)$$

$$RMSE(RUL, \widehat{RUL}) = \left(\frac{1}{n}\sum_{i=1}^{n}(RUL_i - \widehat{RUL}_i)^2\right)^{1/2} \quad (5)$$

$$NASA(RUL, \widehat{RUL}) = 0.5\, RMSE + 0.5\, s_c \quad (6)$$

in which $\alpha$ is the overestimation penalty equal to 1/13 if the RUL is underestimated (i.e., $\widehat{RUL}_i < RUL_i$) and equal to 1/10 for overestimations. Note that we can substitute these values and rewrite Eq. 4 as follows using an indicator function, which will allow for easier implementation with automatic differentiation coding environments:

$$u = \left(\frac{1}{13} + \frac{3}{130}\mathbf{1}_{\widehat{RUL}_i > RUL_i}\right)|RUL_i - \widehat{RUL}_i| \quad (7)$$

$$s_c(RUL, \widehat{RUL}) = \frac{1}{n}\sum_{i=1}^{n}\exp(u) - 1 \quad (8)$$

This alternative formulation essentially "upgrades" the $\alpha$ penalty from a base value of 1/13 to 1/10 when RUL is overestimated. However, the loss function in Eq. 6 only considers RUL errors. We propose utilizing the binary cross-entropy loss function $BCE(\mathbf{y}, \widehat{\mathbf{y}})$ for $q$ classification outputs. Finally, we obtain the overall loss function by weighing the terms introduced above:

$$L(\mathbf{y}, \widehat{\mathbf{y}}) = NASA(RUL, \widehat{RUL}) + \gamma BCE(\mathbf{y}, \widehat{\mathbf{y}}) \quad (9)$$

where $\gamma$ is a tunable weight attached to the classification loss term. Because the NASA score is based on RUL regression, we expect this term will dominate the overall loss value due to the greater magnitude compared to the BCE. For this application, we set $\gamma = 10$ to balance the regression and classification objectives.

Additionally, because benchmarking comparisons between multiple machine learning regressors have not yet been provided in previous work on the N-CMAPSS dataset (DeVol et al., 2021; Lövberg, 2021; Solís-Martín et al., 2021), we aim to compare the performance of four state-of-the-art machine learning methods: random forests (RFs) (Genuer et al., 2017), extreme random forests (ERFs) (also known as extra trees) (Maier et al., 2015), XGBoost (XGB) (Chen & Guestrin, 2016) and artificial neural networks (ANNs) (Mahamad et al., 2010). Specifically, we will compare the performance of the tree-based ensemble regressors trained to minimize mean squared error (MSE) to an ANN that minimizes our proposed loss function in Eq. 9. For completeness, we will also compare these results to an ANN minimizing MSE. To evaluate the quality of classification predictions, we will use the area under receiver operating characteristics (AUROC) and precision-recall curves (AUPR) metrics to evaluate performance at all possible thresholds. Meanwhile, root-mean-square error (RMSE), the NASA scoring function detailed in Eq. 6, mean absolute error (MAE), and MAE normalized as a percentage of the unit's lifetime will be reported for judging regression quality for RUL predictions.

## 3. RESULTS

For reproducibility, results will be reported using the built-in N-CMAPSS dataset split, as in (DeVol et al., 2021). This dataset split is notable for having a testing set with entire engine units that are unseen in the training set, making the benchmark problem more realistic and challenging. The split follows an approximately 60%-40% training-testing ratio, with 4089 cycles in the training set and 2736 cycles in the test set. Following the feature extraction method detailed in Section 2.2, we extract 129 features. Using the min-max standardization and PCA orthogonalization methods from the popular scikit-learn package (Pedregosa et al., 2012), we normalize the training set and apply these learned transformations to the testing set.

To prepare the regressors minimizing the MSE loss function, it is necessary to scale the labels such that the RUL regression error does not dominate the MSE calculation. This is done by simply multiplying the binary encoded labels in $\mathbf{y}$ by 100, thereby putting the binary labels in the same magnitudes as the RUL labels. After training the models, the AUROC and AUPR metrics are then computed using the resulting classification predictions on the test set in $\widehat{\mathbf{y}}$ to serve as robust indicators of performance averaged across all possible thresholds. Table 3 shows the classification and regression results for RF, ERF, XGB, an ANN also trained on MSE (ANN-MSE), and an ANN trained on the custom loss function derived in Section 2.5 (ANN-Flux). Results with and without the PCA orthogonalization step are included, demonstrating the impact of the preprocessing procedure on minimally tuned models. The RF and ERF regressors, implemented using scikit-learn, each contain 100 base estimators. The XGBoost method also uses 100 estimators, with the learning rate $\eta$ set to 0.3 and the max depth of a tree set to 6. Both ANNs share the same shallow architecture of two hidden layers with 64 and 32 neurons each with RELU activations, employing the ADAM optimizer and trained for 5000 epochs. All methods are implemented using the Julia 1.7.3 programming language (Bezanson et al., 2014) and both ANNs are designed using the Flux deep learning backend, which allows for auto-differentiation of custom loss functions (Innes, 2018). The results in Table 3 may be further improved with hyperparameter optimization techniques and merely illustrate the potential for the simultaneous prediction of eventual failures alongside RUL.

| Output | Validation Metric | RF | + PCA | ERF | + PCA | XGB | + PCA | ANN-MSE | + PCA | ANN-Flux | + PCA |
|---|---|---|---|---|---|---|---|---|---|---|---|
| Health State | AUROC | 0.99 | 0.99 | 0.99 | 0.99 | 0.99 | 0.99 | 0.99 | 0.99 | 0.99 | 0.99 |
| | AUPR | 0.99 | 0.98 | 0.99 | 0.99 | 0.99 | 0.98 | 0.99 | 0.99 | 0.99 | 0.99 |
| Fan Failure | AUROC | 0.81 | 0.92 | 0.79 | 0.91 | 0.88 | 0.94 | 0.82 | 0.97 | 0.87 | 0.99 |
| | AUPR | 0.66 | 0.88 | 0.62 | 0.87 | 0.79 | 0.91 | 0.74 | 0.92 | 0.82 | 0.99 |
| LPC Failure | AUROC | 0.81 | 0.91 | 0.80 | 0.91 | 0.90 | 0.93 | 0.75 | 0.96 | 0.82 | 0.97 |
| | AUPR | 0.67 | 0.83 | 0.66 | 0.81 | 0.84 | 0.89 | 0.61 | 0.93 | 0.70 | 0.95 |
| HPC Failure | AUROC | 0.84 | 0.93 | 0.84 | 0.93 | 0.94 | 0.96 | 0.73 | 0.99 | 0.84 | 0.99 |
| | AUPR | 0.81 | 0.91 | 0.81 | 0.90 | 0.94 | 0.95 | 0.70 | 0.99 | 0.82 | 0.99 |
| HPT Failure | AUROC | 0.82 | 0.89 | 0.81 | 0.88 | 0.89 | 0.90 | 0.91 | 0.96 | 0.91 | 0.97 |
| | AUPR | 0.79 | 0.87 | 0.79 | 0.85 | 0.88 | 0.90 | 0.92 | 0.94 | 0.92 | 0.97 |
| LPT Failure | AUROC | 0.80 | 0.88 | 0.80 | 0.88 | 0.87 | 0.91 | 0.83 | 0.90 | 0.83 | **0.95** |
| | AUPR | 0.78 | 0.86 | 0.79 | 0.85 | 0.87 | 0.90 | 0.85 | 0.89 | 0.85 | **0.95** |
| RUL | RMSE | 10.14 | 10.72 | 10.19 | 10.16 | 9.64 | 10.13 | 9.23 | 9.51 | 8.27 | **7.75** |
| | NASA | 5.81 | 6.20 | 5.85 | 5.84 | 5.50 | 5.82 | 5.25 | 6.02 | 4.61 | **4.34** |
| | MAE (cycles) | 7.94 | 8.59 | 8.02 | 8.10 | 7.52 | 7.77 | 7.23 | 7.29 | 6.14 | **5.87** |
| | MAE (%) | 10.72 | 11.51 | 10.89 | 10.85 | 10.12 | 10.45 | 9.66 | 9.53 | 8.07 | **7.72** |

Table 3. Classification and regression results for N-CMAPSS dataset, with the proposed PCA-orthogonalized ANN-Flux method generally outperforming the other benchmark methods

Benchmarked on a single machine running with an Intel Core i7-10750H CPU @ 2.60 GHz processor with 32 GB RAM, it takes approximately just 60 seconds in total to train and evaluate the RF, ERF, and XGB methods. On the same processor, the Flux models take approximately 250 seconds to train with a NVIDIA GeForce RTX 2070 Super GPU. The feature extraction is the longest step in terms of runtime, taking ~450 seconds to load the dataset and extract all 129 features for the training and testing sets.

The ANN-Flux method with the PCA preprocessing step accurately predicts the current health state and the eventual failure component(s) with AUROC and AUPR scores exceeding 0.95 for each output. This is especially notable considering the significant overlap of the failing components depending on the failure modes (see Table 1). In addition, the RUL prediction also outperforms the other techniques tested. The parity plot in Figure 3a) visualizes the ANN-Flux RUL predictions in the testing set versus the ground truth labels. In addition, producing a figure similar to (DeVol et al., 2021),

Figure 3b) also illustrates the ANN-Flux RUL predictions with the ground truth RUL labels sorted from least-to-greatest. Compared to previously published efforts on the N-CMAPSS benchmark on the NASA split, the RUL prediction is significantly improved. Our RMSE of 7.75 and NASA score (smaller is better) of 4.34 compare favorably with the RMSE of 12.50 and NASA score of 7.50 in previously published work—a 38% and 42% reduction, respectively (DeVol et al., 2021).

We note that these RUL predictions are directly output from the ANN-Flux model and further considerations may improve their quality and usefulness in practice. For example, despite the asymmetric NASA scoring function favoring conservative underestimates, the average prediction error still slightly overestimates the ground truth by 0.65 cycles. This contrasts with the training prediction error, which on average underestimates the ground truth by 0.50 cycles. Further adjustments on the overestimation penalty $\alpha$ and the classification loss weight $\gamma$ may skew the prediction error

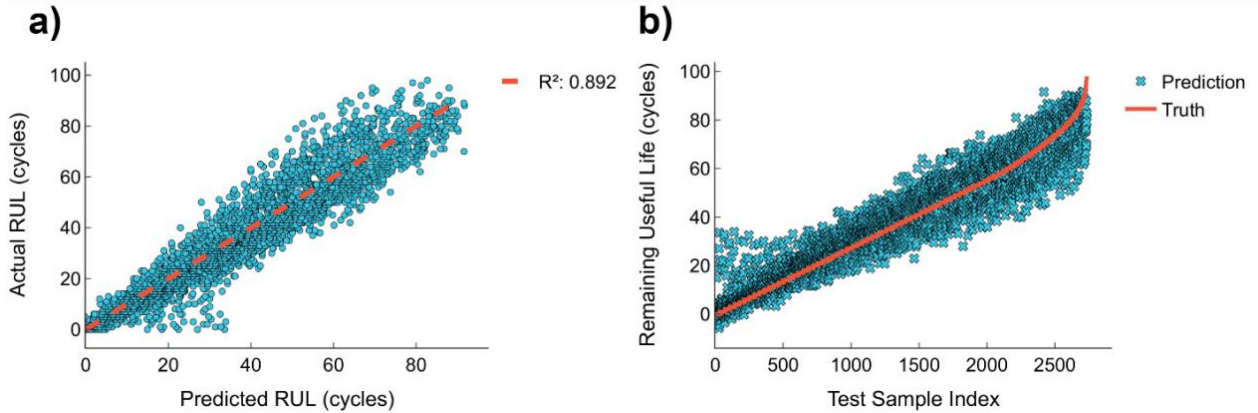

Figure 3. **a)** Parity plot comparing actual and predicted RUL values for ANN-Flux predictions on the N-CMAPSS testing engine unit set; **b)** ANN-Flux predictions scatter with ground truth labels sorted from least-to-greatest

towards underestimation. In addition, we have not instituted hard constraints to guarantee nonnegative RUL values. Post-processing transformations such as the ReLU function can be implemented in the future to rectify the outputs such that all resulting RUL predictions are nonnegative.

Notably, the PCA orthogonalization pre-processing step has a profound impact on classification performance for the eventual failure of the mechanical components. This is especially true for ANN-MSE, which has AUROC and AUPR scores increasing by at least 0.20 for compressor failures when PCA orthogonalization is performed prior to classification. These findings are consistent among all attempted machine learning methods. However, PCA orthogonalization did not appear to improve the regression performance in the same way; 3 out of the 5 attempted methods had increased RMSE when inputs were orthogonalized. This suggests that using PCA to orthogonalize these extracted features is especially useful for binary classification predictions, but may not always lead to better results for minimizing RUL error.

Having additional classification outputs enables explainable analysis of RUL predictions along various slices of the dataset. For example, Figure 4 illustrates the RUL prediction errors for unhealthy versus healthy cycles. Intuitively, the interquartile range for unhealthy operating cycles is substantially narrower, indicating that RUL predictions on average improve throughout the life of the engine unit.

It is also useful to determine whether there are certain components with higher variance in RUL prediction errors; by observing the RUL prediction errors on a per-component basis, operators can glean more information and make targeted decisions based on their confidence of the prognosis. Similar to Figure 4, Figure 5 plots the RUL prediction error spread of the test set for each of the labeled eventual mechanical component failures. Figure 5 demonstrates that the RUL prediction errors have a median centered near 0 for each mechanical component and there is no significant component-based bias identified. Relatively, the compressor failures have a tighter concentration around 0 and the turbine failures are more negatively skewed, indicating more underestimates, but we note that it is difficult to draw definitive conclusions due to overlapping failures.

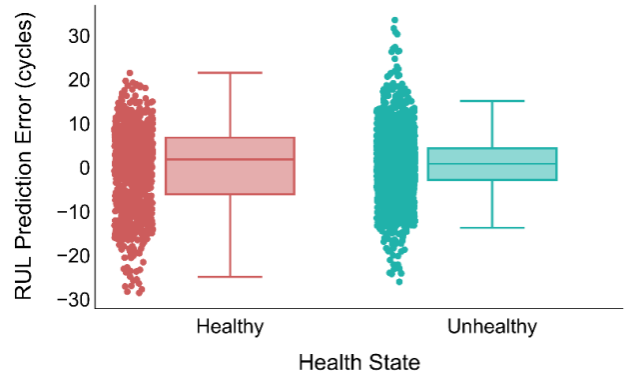

Figure 4. Box-and-whisker plot for RUL error for healthy vs unhealthy operation cycles

**4. DISCUSSION**

Our findings have broad economic implications beyond engine prognostics, as a similar approach could potentially be applied for other PHM applications. Our approach is enabled by expanding the formulation of the 2021 PHM Data Challenge to include these objectives, taking full advantage of the provided labels in the N-CMAPSS dataset. Previous research on this dataset also utilized the labeled health state as an input to improve RUL predictions [9]; we have relaxed assumptions by instead learning the health state as an output.

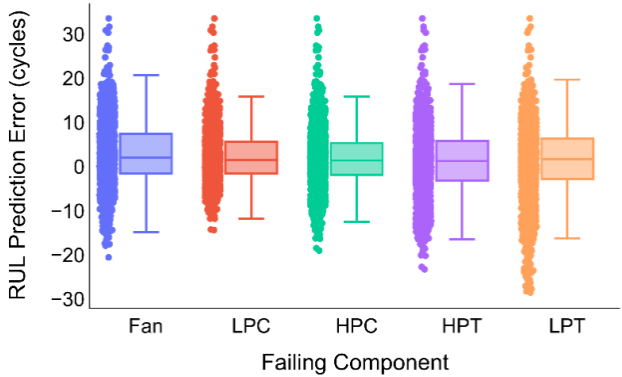

Figure 5. Box-and-whisker plots for RUL prediction errors for each eventual failing component

The computation effort of our approach compared to past work is also noteworthy. Table 4 compares the number of trainable parameters in ANN-Flux compared to the 2021 PHM Data Challenge winners' reported number of parameters. Lövberg does not report the total number of trainable parameters, but utilizes a four-layer ANN for normalization as a pre-processing step before a 10-layer CNN with dilated convolutions made for RUL predictions [9]. ANN-Flux is remarkably simple, with number of parameters approximately two orders of magnitude fewer compared to the deep CNNs of prior work. In a realistic scenario with larger datasets, smaller networks are less expensive to run in real-time, streamlining inferencing efforts. Although our method requires hand-selected features prior to training an ANN, the extracted features are simple statistical features and do not require significant domain expertise. Perhaps surprisingly, predicting RUL in addition to the eventual failure component(s) and current health state does not appear to negatively impact the prediction error, as our approach results in a RMSE reduction of 38% for the same split of the dataset [10].

|  | **Lövberg (2021)** | **DeVol *et al.* (2021)** | **Solís-Martín *et al.* (2021)** | **ANN-Flux** |
|---|---|---|---|---|
| # Trainable Parameters | N/A | 1,030,000 | 4,089,465 | 10,631 |

Table 4. Number of trainable parameters compared with 2021 PHM Data Challenge winners

To the authors' knowledge, our work is also the first to compare multiple state-of-the-art regression approaches for predicting component failures and RUL estimation for the N-CMAPSS benchmark. We also provide comparisons with and without PCA orthogonalization and for multiple loss functions, with tangible improvements for both RUL and binary classifications with these computational approaches. We hope that our contributions encourage future benchmarking efforts on the N-CMAPSS dataset and for PHM research.

Despite these advancements, important limitations remain that require addressing in future work. Firstly, while RUL prediction is improved over past work, the prediction errors still have a large variance. Integration with physics is suggested in the future to improve the confidence of RUL predictions. In addition, failure data are difficult to obtain in practice, and as a result, industrial datasets are often imbalanced (Santos et al., 2018), threatening the utility of fully supervised learning techniques. As a result, more research is required in semi-supervised and unsupervised methods to at least lighten the supervision requirement for AI algorithms to provide accurate prognoses. In addition, while PCA orthogonalization vastly improved the component failure predictions, the derived PC variables lack physical meaning, hindering the explainability of the input features. This step makes the current formulation incompatible with explainable AI (XAI) methods such as SHAP, which attempt to explain black-box model predictions in terms of additive marginal contributions of features (Senoner et al., 2021). While being able to accurately isolate eventual failures on a component-level provides inherent explainability compared to previous efforts, we leave XAI integration for future work.

## 5. CONCLUSION

Our work as benchmarked on the N-CMAPSS dataset uniquely demonstrates the potential for an approach that simultaneously detects the current health state, predicts which component(s) will fail, and then estimates the number of cycles until failure. In essence, this integrates the important disciplines of anomaly detection and fault diagnosis—conventionally requiring multiple models—in one prognostic model that makes accurate predictions, even for presently healthy units. Our main contributions and findings for this research effort are restated as follows:

- Reformulated and expanded the scope of the 2021 PHM Data Challenge to include health state detection and eventual failure prognosis;
- Customized loss function derived to simultaneously balance classification and regression objectives;
- Accurately predicted health state and eventual failures, with AUROC and AUPR exceeding 0.95 for each classification prediction accomplished with the ANN-Flux methodology;
- RUL RMSE reduced by 38% for the same dataset split and with less computational effort required for training compared to prior work.

The authors hope that these contributions will help bolster PHM research and Industry 4.0 efforts to improve safety, lower costs, and enhance decision-making in the age of Big Data.

## 6. DATA AVAILABILITY

We plan on making all code for this paper fully available on GitHub for maximum transparency and encourage reproducibility to further N-CMAPSS as a benchmark for PHM research. The N-CMAPSS dataset is publicly available for download in NASA's Prognostics Center of Excellence Data Repository: https://www.nasa.gov/content/prognostics-center-of-excellence-data-set-repository.

## ACKNOWLEDGEMENT

The authors would like to thank the S. M. Wu Manufacturing Research Center as well as the Uncertainty Quantification & Scientific Machine Learning Group at the University of Michigan for invaluable feedback and support.

## REFERENCES

Bagavathiappan, S., Lahiri, B. B., Saravanan, T., Philip, J., & Jayakumar, T. (2013). Infrared thermography for condition monitoring – A review. *Infrared Physics & Technology*, *60*, 35–55. https://doi.org/10.1016/J.INFRARED.2013.03.006

Bezanson, J., Edelman, A., Karpinski, S., & Shah, V. B. (2014). Julia: A Fresh Approach to Numerical Computing. *SIAM Review*, *59*(1), 65–98. https://arxiv.org/abs/1411.1607v4

Bousdekis, A., Papageorgiou, N., Magoutas, B., Apostolou, D., & Mentzas, G. (2017). A Proactive Event-driven Decision Model for Joint Equipment Predictive Maintenance and Spare Parts Inventory Optimization. *Procedia CIRP*, *59*, 184–189. https://doi.org/10.1016/J.PROCIR.2016.09.015

Chao, M. A., Kulkarni, C., Goebel, K., & Fink, O. (2021). *PHM Society Data Challenge 2021*. *June*, 1–6.

Chao, M., Kulkarni, C., Goebel, K., & Fink, O. (2021). Aircraft Engine Run-To-Failure Dataset Under Real Flight Conditions. *NASA Ames Prognostics Data Repository*. https://ti.arc.nasa.gov/tech/dash/groups/pcoe/prognostic-data-repository/#turbofan-2

Chen, T., & Guestrin, C. (2016). XGBoost: A Scalable Tree Boosting System. *Proceedings of the 22nd ACM SIGKDD International Conference on Knowledge Discovery and Data Mining*, 785–794. https://doi.org/10.1145/2939672

Coble, J., Ramuhalli, P., Bond, L., Hines, J. W., & Upadhyaya, B. (2015). A review of prognostics and health management applications in nuclear power plants. *International Journal of Prognostics and Health Management*, *6*(SP3), 1–22.

DeVol, N., Saldana, C., & Fu, K. (2021). Inception Based Deep Convolutional Neural Network for Remaining Useful Life Estimation of Turbofan Engines. *Annual Conference of the PHM Society*, *13*(1). https://doi.org/10.36001/phmconf.2021.v13i1.3109

Genuer, R., Poggi, J. M., Tuleau-Malot, C., & Villa-Vialaneix, N. (2017). Random Forests for Big Data. *Big Data Research*, *9*, 28–46. https://doi.org/10.1016/j.bdr.2017.07.003

Innes, M. (2018). Flux: Elegant machine learning with Julia. *Journal of Open Source Software*, *3*(25). https://doi.org/10.21105/JOSS.00602

Jollife, I. T., & Cadima, J. (2016). Principal component analysis: a review and recent developments. *Philos Trans A Math Phys Eng Sci*, *374*(2065). https://doi.org/10.1098/RSTA.2015.0202

Lee, J., Wu, F., Zhao, W., Ghaffari, M., Liao, L., & Siegel, D. (2014). Prognostics and health management design for rotary machinery systems - Reviews, methodology and applications. *Mechanical Systems and Signal Processing*, *42*(1–2), 314–334. https://doi.org/10.1016/j.ymssp.2013.06.004

Li, C., Sanchez, R. V., Zurita, G., Cerrada, M., Cabrera, D., & Vásquez, R. E. (2016). Gearbox fault diagnosis based on deep random forest fusion of acoustic and vibratory signals. *Mechanical Systems and Signal Processing*, *76–77*, 283–293. https://doi.org/10.1016/J.YMSSP.2016.02.007

Liao, L., & Köttig, F. (2014). Review of hybrid prognostics approaches for remaining useful life prediction of engineered systems, and an application to battery life prediction. *IEEE Transactions on Reliability*, *63*(1), 191–207. https://doi.org/10.1109/TR.2014.2299152

Lövberg, A. (2021). Remaining Useful Life Prediction of Aircraft Engines with Variable Length Input Sequences. *Annual Conference of the PHM Society*, *13*(1). https://doi.org/10.36001/PHMCONF.2021.V13I1.3108

Mahamad, A. K., Saon, S., & Hiyama, T. (2010). Predicting remaining useful life of rotating machinery based artificial neural network. *Computers and Mathematics with Applications*, *60*(4), 1078–1087. https://doi.org/10.1016/J.CAMWA.2010.03.065

Maier, O., Wilms, M., von der Gablentz, J., Krämer, U. M., Münte, T. F., & Handels, H. (2015). Extra Tree forests for sub-acute ischemic stroke lesion segmentation in MR sequences. *Journal of Neuroscience Methods*, *240*, 89–100. https://doi.org/10.1016/j.jneumeth.2014.11.011

Pedregosa, F., Varoquaux, G., Gramfort, A., Michel, V., Thirion, B., Grisel, O., Blondel, M., Müller, A., Nothman, J., Louppe, G., Prettenhofer, P., Weiss, R., Dubourg, V., Vanderplas, J., Passos, A., Cournapeau, D., Brucher, M., Perrot, M., & Duchesnay, É. (2012). Scikit-learn: Machine Learning in Python. *Journal of*

*Machine Learning Research*, *12*, 2825–2830. https://arxiv.org/abs/1201.0490v4

Santos, P., Maudes, J., & Bustillo, A. (2018). Identifying maximum imbalance in datasets for fault diagnosis of gearboxes. *Journal of Intelligent Manufacturing*, *29*(2), 333–351. https://doi.org/10.1007/S10845-015-1110-0

Senoner, J., Netland, T., & Feuerriegel, S. (2021). Using Explainable Artificial Intelligence to Improve Process Quality: Evidence from Semiconductor Manufacturing. *Https://Doi.Org/10.1287/Mnsc.2021.4190*, *68*(8), 5704–5723. https://doi.org/10.1287/MNSC.2021.4190

Shao, H., Jiang, H., Lin, Y., & Li, X. (2018). A novel method for intelligent fault diagnosis of rolling bearings using ensemble deep auto-encoders. *Mechanical Systems and Signal Processing*, *102*, 278–297. https://doi.org/10.1016/J.YMSSP.2017.09.026

Solís-Martín, D., Galán-Páez, J., & Borrego-Díaz, J. (2021). A Stacked Deep Convolutional Neural Network to Predict the Remaining Useful Life of a Turbofan Engine. *Annual Conference of the PHM Society*, *13*(1). https://doi.org/10.36001/PHMCONF.2021.V13I1.3110

Tsui, K. L., Chen, N., Zhou, Q., Hai, Y., & Wang, W. (2015). Prognostics and health management: A review on data driven approaches. *Mathematical Problems in Engineering*, *2015*. https://doi.org/10.1155/2015/793161